\documentclass{article}
\usepackage{spconf,amsmath,graphicx}
\usepackage{threeparttable}
\usepackage{amssymb}
\usepackage{amsmath,epsfig}
\usepackage{mathptmx,mathrsfs}
\usepackage{graphicx}
\usepackage{tabularx,colortbl,multirow,float}
\usepackage{enumitem}
\usepackage[hyphens]{url}
\usepackage[colorlinks, urlcolor=black]{hyperref}
\usepackage[caption=false,font=footnotesize]{subfig}
\usepackage{flushend}
\usepackage{changepage}
\usepackage[table]{xcolor}
\usepackage{color,soul}
\usepackage{comment}
\graphicspath{{figs/}}

\usepackage[numbers,sort&compress]{natbib}

\begin{document}
\twocolumn[{%
\vspace{30mm}
{ \large
\begin{itemize}[leftmargin=2.5cm, align=parleft, labelsep=2.0cm, itemsep=4ex,]

\item[\textbf{Citation}]{G. Kwon*, M. Prabhushankar*, D. Temel, and G. AlRegib , "Distorted Representation Space Characterization Through Backpropagated Gradients," 2019 26th IEEE International Conference on Image Processing (ICIP), Taipei, Taiwan, 2019. (*: equal contribution)}

\item[\textbf{Review}]{Date of Publication: September 22, 2019}

\item[\textbf{Codes}]{\url{https://github.com/olivesgatech}}

\item[\textbf{Bib}]  {@INPROCEEDINGS\{Kwon2019\_ICIP,\\
author=\{G. Kwon and M. Prabhushankar and D. Temel and G. AlRegib\},\\ 
booktitle=\{2019 26th IEEE International Conference on Image Processing (ICIP)\},\\
title=\{Distorted Representation Space Characterization Through Backpropagated Gradients\},\\
year=\{2019\}, \}
}

\item[\textbf{Copyright}]{\textcopyright 2019 IEEE. Personal use of this material is permitted. Permission from IEEE must be obtained for all other uses, in any current or future media, including reprinting/republishing this material for advertising or promotional purposes,
creating new collective works, for resale or redistribution to servers or lists, or reuse of any copyrighted component
of this work in other works. }


\item[\textbf{Contact}]{
\{gukyeong.kwon, mohit.p, alregib\}@gatech.edu\\
\url{http://ghassanalregib.com/}\\}
\end{itemize}
\thispagestyle{empty}
\newpage
\clearpage
\setcounter{page}{1}}}]

\title{Distorted Representation Space Characterization Through Backpropagated Gradients}
%
\name{Gukyeong Kwon*\thanks{*Equal contribution}, Mohit Prabhushankar*, Dogancan Temel, and Ghassan AlRegib}
\address{Center for Signal and Information Processing,\\ School of Electrical and Computer Engineering,\\ Georgia Institute of Technology, Atlanta, GA, 30332-0250\\ \{gukyeong.kwon, mohit.p, cantemel, alregib\}@gatech.edu}
%
%
%
\ninept
\maketitle
\begin{abstract}
In this paper, we utilize weight gradients from backpropagation to characterize the representation space learned by deep learning algorithms. We demonstrate the utility of such gradients in applications including perceptual image quality assessment and out-of-distribution classification. The applications are chosen to validate the effectiveness of gradients as features when the test image distribution is distorted from the train image distribution. In both applications, the proposed gradient based features outperform activation features. In image quality assessment, the proposed approach is compared with other state of the art approaches and is generally the top performing method on TID 2013 and MULTI-LIVE databases in terms of accuracy, consistency, linearity, and monotonic behavior. Finally, we analyze the effect of regularization on gradients using CURE-TSR dataset for out-of-distribution classification. 
\end{abstract}
\vspace{-1.5mm}
\section{Introduction}
\label{sec:intro}
\vspace{-1.5mm}
The recent ubiquitous deployment of deep learning algorithms across multiple vision applications can be attributed to the statistical commonalities present in all image data~\cite{Goodfellow2016, Bengio2013}. Feedforward neural networks learn nonlinear transformations with the objective of obtaining useful representations of complex inputs. These representations make the subsequent applications easier. For instance, in image recognition, the authors in~\cite{krizhevsky2012imagenet} successfully apply a linear classifier on a representation space that is obtained by multiple weight parameters spread across layers. The high accuracy is achieved because of the constancy of the train and test domain statistics that allows the weights to transform the test images into the trained representation space. While the concept of domain transformation is not new~\cite{stockham1972image}, the advent of backpropagation algorithm~\cite{rumelhart1986learning} has allowed the learning of representation spaces using large data. During training, the backpropagation adjusts the weights between layers to minimize the difference between obtained output and desired output. The measure of adjustment is controlled by optimization procedures. Generally, regularized gradient based optimization procedures are adopted to obtain the direction of weight adjustment. During testing, images are projected onto the learned representation space. If there is no change in the train and the test domain, the representation space is optimized to perform the task it was trained for.

Domain adaptation techniques have been been proposed to bridge the distributional gap, when there are differences in the train and test image statistics~\cite{patel2015visual}. These techniques essentially re-train the weight parameters to adjust the representation space to include the patterned shift of the test domain. Once the training is done, the test images are again projected onto the re-trained weights to presumably lie on the optimized representation space. However, in the cases where there is not enough data in the test domain for the representation space to adapt, domain adaptation techniques cannot be applied. Similarly, when the domain shift is caused by various types and levels of distortions such as noise which does not possess statistical commonalities with natural images, domain adaptation techniques may not perform as desired. Applications that satisfy these criteria include image quality assessment and out-of-distribution classification. 

In this paper, we tackle the case where there is a considerable shift between the train and test domains because of distortions and where there is not enough test data for the weights to adapt to the test distribution. The contributions of this paper are threefold:
\begin{itemize}
    \item We introduce the concept of using directional information provided by gradients as a feature.
    \item We use gradients as error projection spaces to quantify perceptual dissimilarity between pristine and distorted images and validate this quantification on image quality assessment.
    \item We analyze the effect of regularization on gradients for out-of-distribution classification. 
\end{itemize}
We use an autoencoder architecture to demonstrate the validity of gradients as features. In Section~\ref{sec:Background}, we set up the autoencoder architecture. The proposed method is introduced in Section~\ref{sec:Framework}. The application of the proposed method along with the conducted experiments are reported in Sections~\ref{sec:Experiments} and~\ref{sec:results}.

\vspace{-1.5mm}
\section{Background}
\label{sec:Background}
\vspace{-1.5mm}
    
\begin{figure}[tb]
 	\centering
 	\includegraphics[width=1\columnwidth]{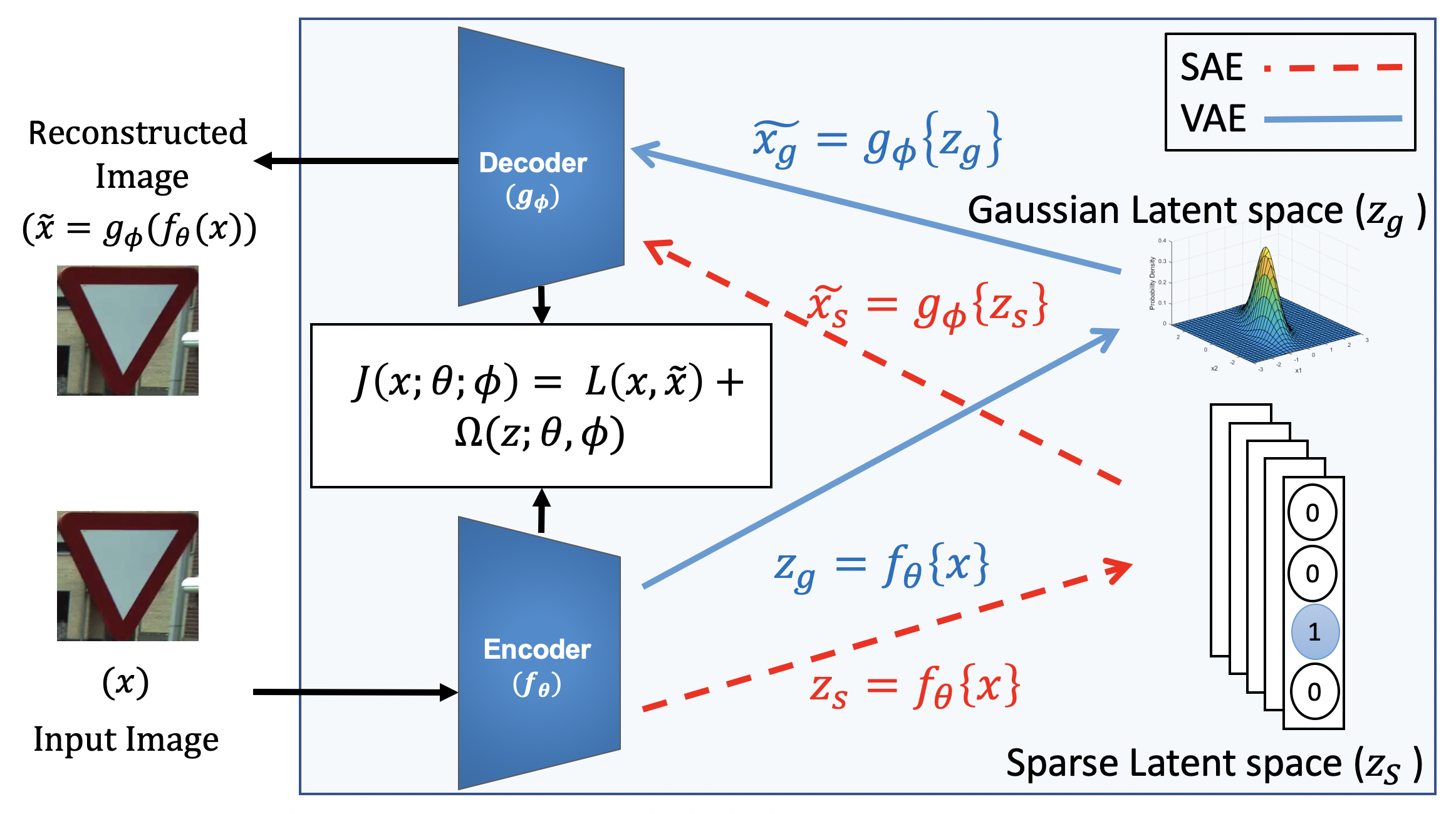}
 	\caption{Block diagram for Variational Autoencoder (VAE) and Sparse Autoencoder (SAE). The image $x$ is taken from the CURE-TSR dataset}\label{fig:AE_BlockDiagram}
 	\vspace{-0.3cm}
\end{figure}

An autoencoder is an unsupervised learning network which learns a regularized representation of inputs to reconstruct them as its output~\cite{Goodfellow2016}. 
The representation space of autoencoders can be constrained to have desirable properties in its latent space. An autoencoder constrained to be sparse in its latent space is called a Sparse Autoencoder (SAE)~\cite{ng2011sparse}. Similarly, an autoencoder which reconstructs input data while constraining the latent space to follow a known distribution such as Gaussian distribution, is called a Variational Autoencoder (VAE)~\cite{Kingma2013}. Both the SAEs and VAEs consist of the same basic architecture as shown in Fig.~\ref{fig:AE_BlockDiagram}. The encoder, $f_\theta$, is parameterized to map inputs $x$ to the latent representation, $z$, and the decoder, $g_\phi$, learns to reconstruct inputs using the latent representation. That is, given input $x \in \mathbb{R}^{H \times W \times C}$, 
\begin{equation}
    z = f_\theta (x) \quad \tilde{x} = g_\phi (z),
\end{equation}
where $H, W, C$ are height, width, channel of input image, respectively. The notations $z_s, \tilde{x_s}$ and $z_g,\tilde{x_g}$ are used to differentiate SAE and VAE in Fig.~\ref{fig:AE_BlockDiagram}. Training is carried out by minimizing the loss function $J(x; \theta, \phi)$ defined as follows:
\begin{equation}\label{eq:loss}
    J(x; \theta, \phi) = \mathcal{L}(x, g_{\phi}(f_{\theta}(x))) + \Omega(z; \theta, \phi),
\vspace{-1.5mm}
\end{equation}
where $\mathcal{L}$ is a reconstruction error which measures the dissimilarity between the input and the reconstructed image and $\Omega$ is a regularization term. 

SAEs differ from VAEs in the loss function $J(x; \theta, \phi)$. For SAEs, we enforce latent space sparsity by using mean square error and elastic net regularization~\cite{zou2005regularization} for $\mathcal{L}$ and $\Omega$, respectively. The cost function for training our SAE is,
\begin{equation}
    J(x;\theta,\phi) = \lVert x -\tilde{x} \rVert_2^2 + \beta\lVert z_s \rVert_1 + \lambda\lVert \theta;\phi \rVert_2^2,
\end{equation}
where, $\theta$ and $\phi$ are the weight parameters for the encoder and decoder, respectively. $\beta=3$ and $\lambda=3e^{-3}$ are set as suggested by the authors in~\cite{ng2012ufldl}. The loss function of VAEs are defined as
\begin{equation}
    J(x;\theta, \phi) = - \mathbb{E}_{g_\phi(z|x)}[\log f_\theta(x | z)] + \text{KL}[g_\phi(z|x)|| f(z)].
\end{equation}
KL is the Kullback Leibler divergence between two distributions and we assume $f(z) = N(z| \textbf{0}, I)$. Thus, KL divergence constrains the latent space of VAEs to be a Gaussian distribution. The first term in the loss corresponds to the reconstruction error, $\mathcal{L}$, and the second term is a latent constraint, $\Omega$ in Eq.~(\ref{eq:loss}). In this paper, we use the trained representation spaces of both SAEs and VAEs to validate our methodology.
\vspace{-1.5mm}

\vspace{-1.5mm}
\section{Gradient Interpretation and Generation}
\label{sec:Framework}
\vspace{-1.5mm}
\begin{figure}[tb]
 	\centering
 	\includegraphics[width=.65\columnwidth]{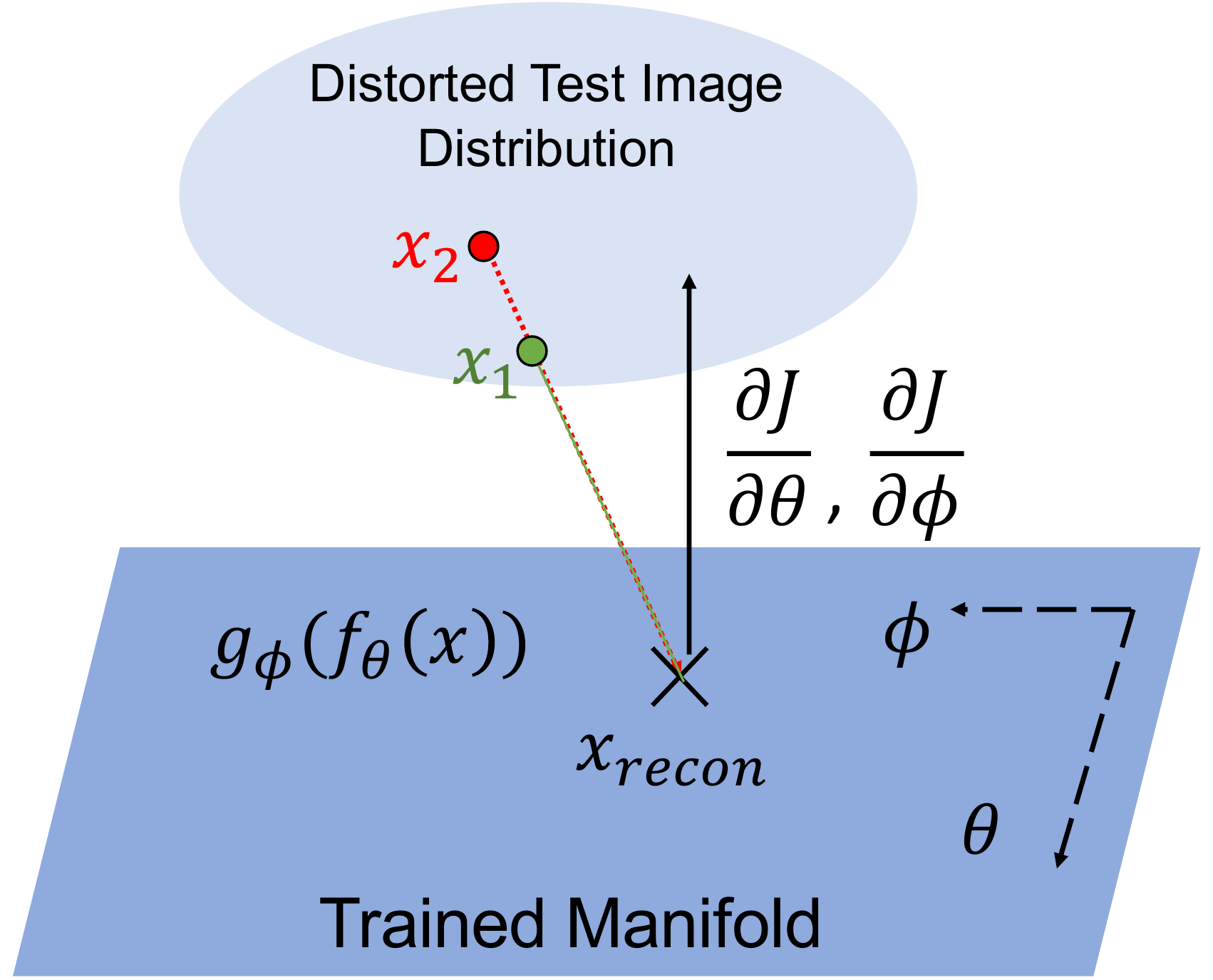}
 	\vspace{-0.2cm}
 	\caption{Geometric visualization for gradients from test images.}\label{fig:manifold}
 	\vspace{-0.2cm}
\end{figure}
\begin{figure}[tb]
 	\centering
 	\includegraphics[width=.99\columnwidth]{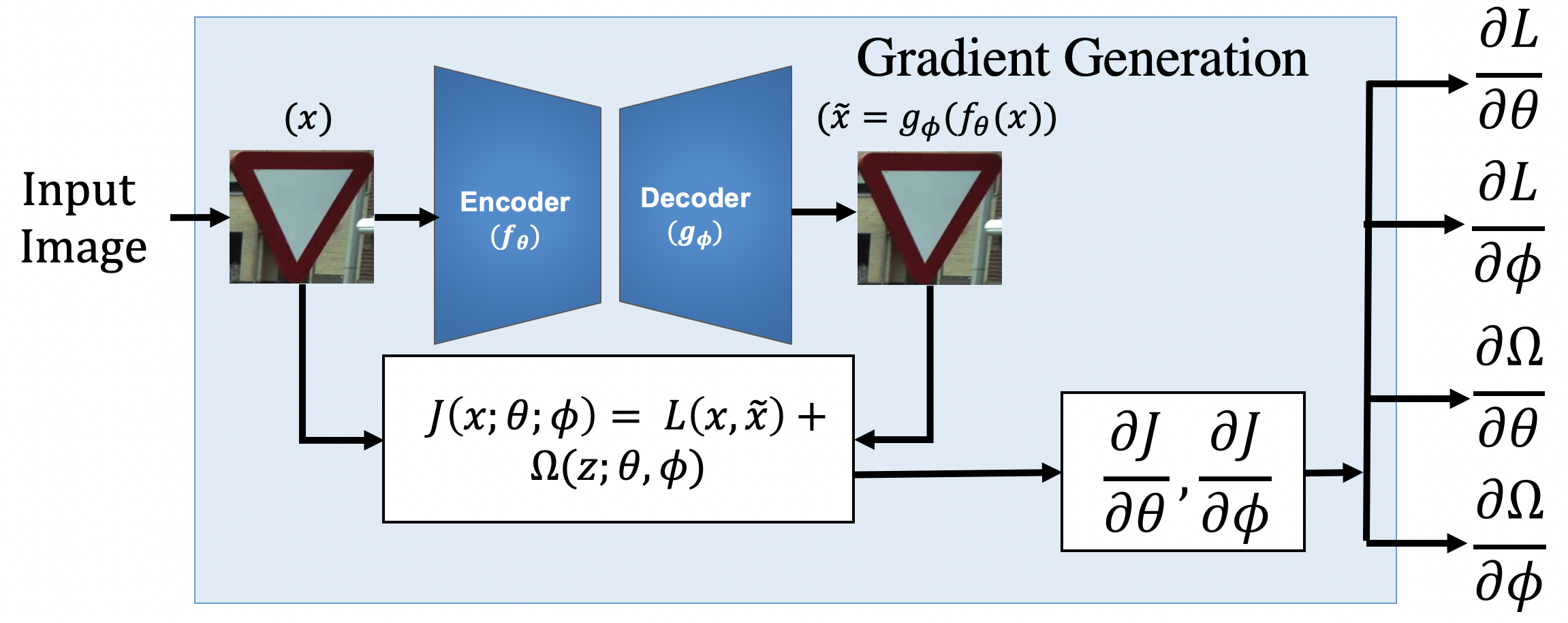}
 	\vspace{-0.2cm}
 	\caption{Block Diagram for gradient generation. The image $x$ is taken from the CURE-TSR dataset.}\label{fig:GradGen}
 	\vspace{-0.2cm}
\end{figure}
The latent representation spaces $z_s$ and $z_g$ are low-dimensional manifolds embedded in a high dimensional input space. The activation of encoder is a projection of image on the learned manifold~\cite{Bengio2013}. The generalizability of autoencoder is ensured when test images are drawn from the same training distribution and are well projected in the vicinity of the learned manifold. However, the presence of distortions in test images causes their activations to move away from the representation manifold. Unless this movement is tangential to the manifold, the representations may not be accurate enough for the subsequent tasks. This limits the capability of activation from the encoder in characterizing the distortions in the input space. This is shown in Fig.~\ref{fig:manifold}, where both $x_1$ and $x_2$ have the same activation $x_{recon}$ on the manifold.

To alleviate this problem, we propose to utilize the gradients of trained autoencoder to characterize distortions in the test images. The geometric interpretation of backpropagated weight gradients is explained in subsection~\ref{subsec:Geometry}, while the gradient generation methodology is described in subsection~\ref{subsec:Gradient Generation}.
\vspace{-1.5mm}
\subsection{Geometric Interpretation}
\label{subsec:Geometry}
\vspace{-1.5mm}
In Fig.~\ref{fig:manifold}, we visualize the geometric interpretation of gradients. We assume the manifold of reconstructed training images to be linear for the simplicity of explanation. As shown in the figure, the manifold of reconstructed training images is considered to be spanned by the weights of the encoder and the decoder. When the test images are the variations of training images, the autoencoder approximates the reconstruction, $x_{recon}$, by projecting them on the learned reconstructed training image manifold, $g_\phi (f_\theta (x))$. The gradient of weights can be calculated using the loss defined in Eq.~(\ref{eq:loss}) and the backpropgataion, $\frac{\partial J}{\partial \theta}, \frac{\partial J}{\partial \phi}$. These gradients represent required changes in the reconstructed image manifold to incorporate the test images. Therefore, we use these gradients to characterize the orthogonal distortions toward the test images. For instance, when $x_1$ and $x_2$ are projected onto different activations in the direction of the weight gradients $\frac{\partial J}{\partial \theta}, \frac{\partial J}{\partial \phi}$, they become distinguishable in the gradient space. We note that the assumption of linear manifold does not hold in most of practical scenarios but the geometric interpretation of gradients is still applicable to nonlinear manifolds. 
\vspace{-1.5mm}
\subsection{Gradient Generation Framework}
\vspace{-1.5mm}
\label{subsec:Gradient Generation}
Fig.~\ref{fig:GradGen} shows the block diagram to generate the required gradients. A test image is passed through a trained network as the input. The feedforward loss from Eq.~\ref{eq:loss} is calculated. This loss is a sum of the reconstruction loss, $\mathcal{L}$, and the regularization loss, $\Omega$. In the loss function, the reconstruction error and the regularization serve different roles during the optimization. Therefore, gradients backpropagated from both terms characterize different aspects of distortions in test image. Hence, we extract and analyze them separately. Both $\mathcal{L}$ and $\Omega$ are differentiated with respect to weight parameters from encoder and decoder, $\theta$ and $\phi$. These four gradients, $\frac{\partial \mathcal{L}}{\partial \theta}, \frac{\partial \mathcal{L}}{\partial \phi}$, $\frac{\partial \Omega}{\partial \theta}$, $\frac{\partial \Omega}{\partial \phi}$, are the generated outputs. Note that the total loss gradient can be calculated as a sum of individual components. In Section~\ref{sec:Experiments}, we show how these generated gradients can be used as features for distorted images.

\vspace{-1.5mm}
\section{Applications and Experiments}
\label{sec:Experiments}
\vspace{-1.5mm}
\begin{figure}[t]
 	\centering
 	\includegraphics[width=\columnwidth]{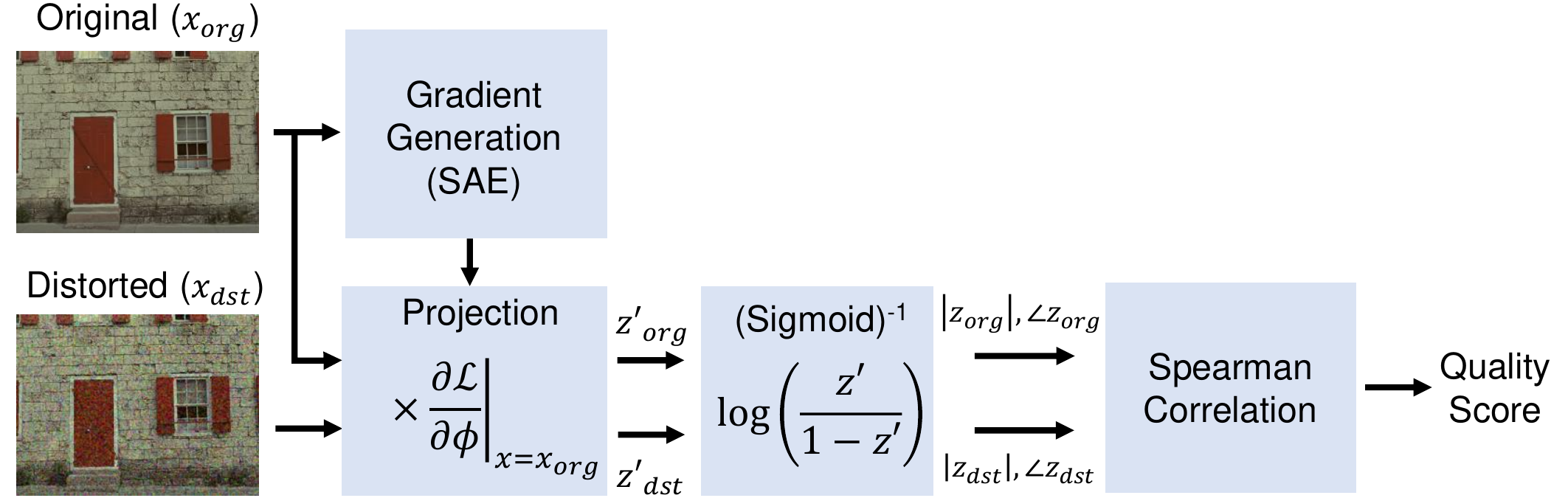}
 	\caption{Block diagram for image quality assessment.}\label{fig:IQA}
 	\vspace{-.5cm}
\end{figure}
Gradients have traditionally been used for providing visual explanations for the decisions made by deep networks~\cite{selvaraju2017grad}. We are not aware of any other work that utilizes gradient information to characterize directional change of learned representation space during test time. Note that the obtained weight gradients are not used to adjust the network parameters. Rather, they are used to characterize the distortions present in test domains. We validate the effectiveness of this characterization using two applications: Out-of-distribution classification (OOD-C) and image quality assessment (IQA).

The weight gradients, $\frac{\partial J}{\partial \theta}, \frac{\partial J}{\partial \phi}$, can be used in two ways. The first approach is to use gradients as error direction indicators by projecting images with complex distortions on the gradient space. The complex distortions can either be a combination of multiple distortions such as distortions generated in the MULTI-LIVE dataset~\cite{jayaraman2012objective} or the human visual system (HVS) specific peculiar distortions such as the ones presented in the TID2013~\cite{ponomarenko2015image} dataset. Hence, this approach of projections onto gradients is explored in the application of IQA. The second approach is to directly use gradients as feature vectors. The assumption here is that the directional change caused by distortions is unique and becomes a characteristic of the distortion. This approach is explored for the application of OOD-C. 

We also analyze the effect of regularization in the characterization of distortions using gradients. We study the gradients backpropagated from regularization and reconstruction errors in the last layer of encoder and decoder. Both these layers serve as the genesis layers for gradients during backpropagated. In particular, we conduct this analysis using VAEs in the application of OOD-C with diverse distortion types. 

\vspace{-1.5mm}
\subsection{Image Quality Assessment}
\label{subsec:IQA}
\vspace{-1.5mm}
Image quality assessment is a classical field of image processing the goal of which is to objectively estimate the perceptual quality of a degraded image. Traditionally, hand-crafted features based on natural scene statistics are extracted from original and distorted images. These features are then mapped to the subjective quality scores~\cite{moorthy2010two, saad2012blind}. Recently, end-to-end deep learning techniques~\cite{kang2014convolutional, temel2016unique} that handle both feature extraction and mapping have been proposed. The authors in~\cite{temel2016unique,Prabhushankar2017_EI} project both original and distorted images onto the same generically trained SAE based representation space to quantify the perceptual difference. A better performance compared to the results in~\cite{temel2016unique} would validate the usage of gradient projections. To facilitate unbiased comparison, we follow the same preprocessing and training steps as described in~\cite{temel2016unique}. The block diagram during testing for the proposed method, is shown in Fig.~\ref{fig:IQA}. Both original and distorted images are projected on the gradient space of decoder. Inversion of nonlinear layer is performed and the Spearman correlation between two projections is calculated to estimate the quality of image.

\vspace{-1.5mm}
\subsection{Out-Of-Distribution Classification}
\label{subsec:OOD-C}
\vspace{-1.5mm}
\begin{figure}[t]
 	\centering
 	\includegraphics[width=\columnwidth]{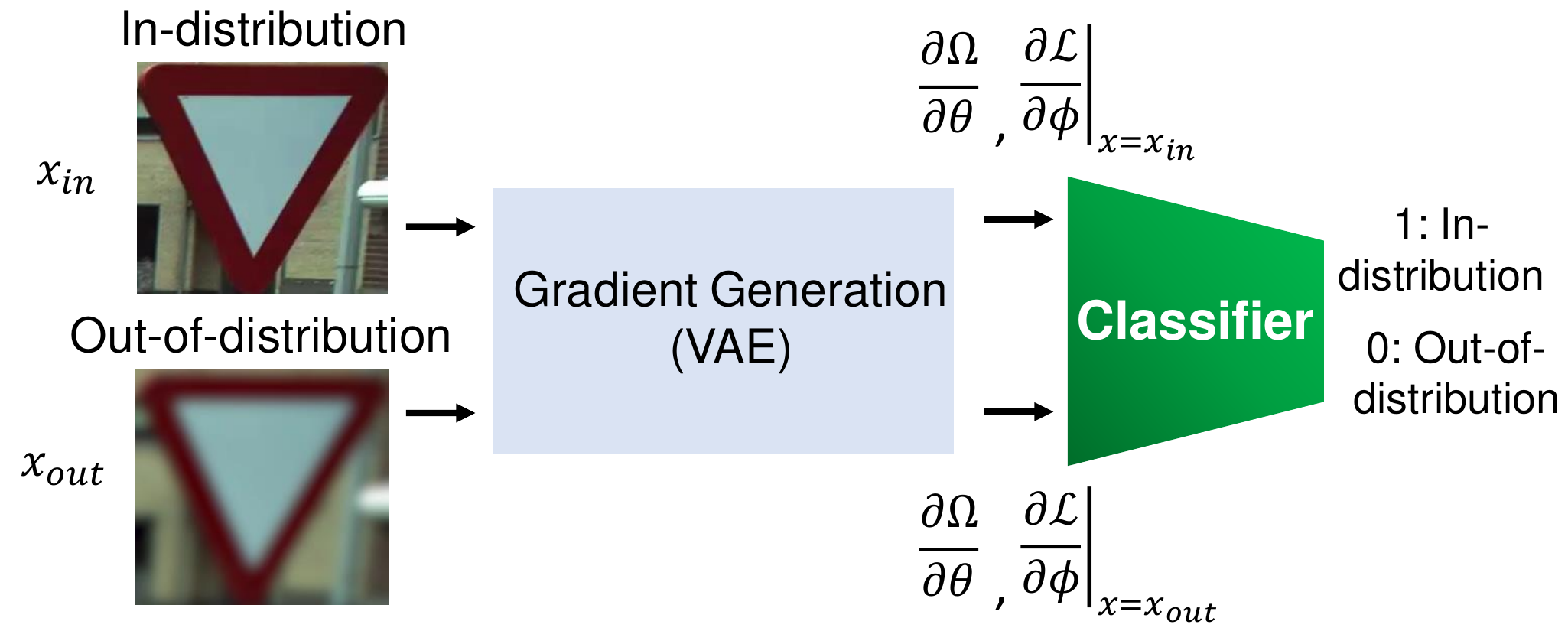}
 	\caption{Block diagram for out-of-distribution classification.}\label{fig:OOD}
 	\vspace{-.5cm}
\end{figure}
Out-of-distribution detection is an emerging research topic to ensure AI safety. The authors in~\cite{Liang2017} propose to use temperature scaling and pre-processing on images to detect OOD images. Zong~\textit{et al.}~\cite{Zong2018} utilize activations and reconstruction error features of autoencoder to perform OOD detection. In~\cite{Fan2018}, a VAE with Gaussian mixture model and a sample energy-based method are utilized to detect OOD. We use the Challenging Unreal and Real Environments for Traffic Sign Recognition (CURE-TSR) dataset~\cite{Temel2017,Temel2018_SPM,Temel2019_VIP} which contains traffic sign images with $12$ different challenging conditions and $5$ challenge levels for out-of-distribution classification. The block diagram of the proposed method is demonstrated in Fig.~\ref{fig:OOD}. We train the VAE using challenge-free training images and generate gradients using challenge-free training images and Gaussian blur training images. The generated gradients are used to classify challenge-free test images (in-distribution) and challenge test images (out-of-distribution) with $7$ challenge types which are decolorization (DE), Codec Error (CE), Noise (NO), Lens Blur (LB), Dirty Lens (DL), and Rain (RA). We compare the performance of classifiers trained with gradient features and activations from the encoder. Also, we analyze the gradient features from the reconstruction error and the regularization in classifying different challenge images.
\vspace{-1.5mm}
\section{Results}
\label{sec:results}
\begin{table*}[htb!]
\tiny

\centering
\caption{Overall performance of image quality estimators.}
\label{tab:tab_results_all}

\begin{threeparttable}

\begin{tabular}{p{0.45cm}p{0.45cm}p{0.45cm}p{0.45cm}p{0.45cm}p{0.45cm}p{0.45cm}p{0.45cm}p{0.45cm}p{0.45cm}p{0.45cm}p{0.45cm}p{0.45cm}p{0.45cm}p{0.45cm}p{0.45cm}p{0.55cm}p{0.80cm}p{1cm}}
\hline


\multirow{3}{*}{{\bf Databases }}                 & \bf PSNR  &\bf PSNR  &\bf SSIM  &\bf MS  &\bf CW &\bf IW  &\bf SR  &\bf FSIM  &\bf FSIMc  &\bf BRIS  & \bf BIQI &\bf BLII &\bf Per &\bf CSV &\bf UNI &\bf COHER &\bf SUMMER &\bf Proposed \\
&\bf HA &\bf HMA & &\bf SSIM  &\bf SSIM  &\bf SSIM  &\bf SIM  & &&\bf QUE & &\bf NDS2 &\bf SIM & & \bf QUE & \bf ENSI & \\ &\cite{ponomarenko2011modified} & \cite{ponomarenko2011modified} &\cite{wang2004image} &\cite{wang2003multiscale} &\cite{sampat2009complex}&\cite{wang2011information}&\cite{zhang2012sr} &\cite{zhang2011fsim} &\cite{zhang2011fsim} &\cite{mittal2012no} &\cite{moorthy2010two} &\cite{saad2012blind}&\cite{temel2015persim} & \cite{Temel201692}&\cite{temel2016unique}  &\cite{Temel_SPIC_2018} &\cite{Temel_SPIC_2018} &
\\ \hline
         
                & \multicolumn{18}{c}{\textbf{Outlier Ratio (OR)}}                           \\ \hline
\textbf{MULTI}  
& 0.013 & 0.009 & 0.016 & 0.013 & 0.093 & 0.013 &\cellcolor{blue!10} \bf 0.000 & 0.018 & 0.016 & 0.067 & 0.024 & 0.078 & 0.004 &\cellcolor{blue!10} \bf 0.000 &\cellcolor{blue!10} \bf 0.000 & 0.031 &\cellcolor{blue!10} \bf 0.000 & \cellcolor{blue!10} \bf 0.000
\\                            
\textbf{TID13}  
&\cellcolor{blue!10} \bf 0.615 & 0.670 & 0.734 & 0.743 & 0.856 & 0.701 & 0.632 & 0.742 & 0.728 & 0.851 & 0.856 & 0.852 & 0.655 & 0.687 & 0.640 & 0.833 & \cellcolor{blue!10} \bf 0.620 & \cellcolor{blue!10} \bf 0.620
 \\ 
 \hline

                & \multicolumn{18}{c}{\textbf{Root Mean Square Error (RMSE)}}                                                                                                                                                                        \\ \hline

\textbf{MULTI}  
& 11.320 & 10.785 & 11.024 & 11.275 & 18.862 & 10.049 &\cellcolor{blue!10} \bf 8.686 & 10.866 & 10.794 & 15.058 & 12.744 & 17.419 & 9.898 & 9.895 & 9.258 & 14.806 & 8.212 & \cellcolor{blue!10} \bf 7.943\\

\textbf{TID13}    & 0.652 & 0.697 & 0.762 & 0.702 & 1.207 & 0.688 &\cellcolor{blue!10} \bf 0.619 & 0.710 & 0.687 & 1.100 & 1.108 & 1.092 & 0.643 & 0.647 & 0.615 & 1.049  & 0.630 & \cellcolor{blue!10} \bf 0.596
 \\ \hline

               & \multicolumn{18}{c}{\textbf{Pearson Linear Correlation Coefficient (PLCC)}}                                                                                                                                                                        \\ \hline

\multirow{2}{*}{{\bf MULTI}}                                  
& 0.801 & 0.821 & 0.813 & 0.803 & 0.380 & 0.847 & 0.888 & 0.818 & 0.821 & 0.605 & 0.739 & 0.389 & 0.852 & 0.852 &  0.872 & 0.622 &\cellcolor{blue!10} \bf  0.901 & \cellcolor{blue!10} \bf 0.908 \\                                          
& -1 & -1 & -1 & -1 & -1 & -1 & 0 & -1 & -1 & -1 & -1 & -1 & -1 & -1 & -1 & -1 & 0 & 

 \\

\multirow{2}{*}{{\bf TID13}}                               
& 0.851 & 0.827 & 0.789 & 0.830 & 0.227 & 0.832 & 0.866 & 0.820 & 0.832 & 0.461 & 0.449 & 0.473 & 0.855 & 0.853 & \cellcolor{blue!10} \bf 0.869 & 0.533  &  0.861 & \cellcolor{blue!10} \bf 0.877\\
& -1 & -1 & -1 & -1 & -1 & -1 & 0 & -1 & -1 & -1 & -1 & -1 & -1 & -1 & 0 & -1 & -1 & 
   \\ \hline

\textbf{}      & \multicolumn{18}{c}{\textbf{Spearman's Rank Correlation Coefficient (SRCC)}}                                                                                                                                                                        \\ \hline

\multirow{2}{*}{{\bf MULTI}}                                          
& 0.715 & 0.743 & 0.860 & 0.836 & 0.631 & \cellcolor{blue!10} \bf 0.884 & 0.867 & 0.864 & 0.867 & 0.598 & 0.611 & 0.386 & 0.818 & 0.849 &  0.867 & 0.554  & \cellcolor{blue!10} \bf 0.884 & \cellcolor{blue!10} \bf 0.887 \\   
& -1 & -1 & 0 & -1 & -1 & 0 & 0 & 0 & 0 & -1 & -1 & -1 & -1 & -1 & 0 & -1 & 0 & 
\\

\multirow{2}{*}{{\bf TID13}}
& 0.847 & 0.817 & 0.742 & 0.786 & 0.563 & 0.778 & 0.807 & 0.802 & 0.851 & 0.414 & 0.393 & 0.396 &0.854 & 0.846 & \cellcolor{blue!10} \bf 0.860 & 0.649  & 0.856 & \cellcolor{blue!10} \bf 0.865 \\

& -1 & -1 & -1 & -1 & -1 & -1 & -1 & -1 & -1 & -1 & -1 & -1 & 0 & -1 & 0 & -1 & 0 &

  \\ \hline

\textbf{}      & \multicolumn{18}{c}{\textbf{Kendall's Rank Correlation Coefficient (KRCC)}}                                                                                                                                                                        \\ \hline

\multirow{2}{*}{{\bf MULTI}}                                              
& 0.532 & 0.559 & 0.669 & 0.644 & 0.457 & \cellcolor{blue!10} \bf 0.702 & 0.678 & 0.673 & 0.677 & 0.420 & 0.440 & 0.268 & 0.624 & 0.655 &  0.679 & 0.399 & 0.698 & \cellcolor{blue!10} \bf 0.702\\                                         

& -1 & -1 & 0 & 0 & -1 & 0 & 0 & 0 & 0 & -1 & -1 & -1 & -1 & 0 & 0 & -1 & 0 & 

 \\  

\multirow{2}{*}{{\bf TID13}}                                          
& 0.666 & 0.630 & 0.559 & 0.605 & 0.404 & 0.598 & 0.641 & 0.629 & 0.667 & 0.286 & 0.270 & 0.277 &\cellcolor{blue!10} \bf 0.678 & 0.654 & 0.667 & 0.474   & 0.667 & \cellcolor{blue!10} \bf 0.677 \\
& 0 & -1 & -1 & -1 & -1 & -1 & -1 & -1 & 0 & -1 & -1 & -1 & 0 & 0 & 0 & -1 & 0 & 
\\ \hline
\end{tabular}
\end{threeparttable}
\vspace{-0.5cm}
\end{table*}
\vspace{-1.5mm}
\begin{table}[tb]
\vspace{-2mm}
\centering
\caption{Out-of-distribution classification accuracy.\label{tab:OOD}}
\begin{tabular}{c|ccc}
\hline\hline
\multicolumn{4}{c}{Accuracy (\%)}                                                        \\ \hline
\multirow{2}{*}{Method}            & \multicolumn{3}{c}{Non-Blur Types}    \\ \cline{2-4} 
                                   & DE             & CE             & NO             \\ \hline
VAE-A $(z_g)$ & 56.51          & 62.01          & 84.72          \\ 
VAE-R
$(\frac{\partial\mathcal{L}}{\partial\mathcal{\phi}})$                                 & \textbf{70.24} & 60.64          & 68.59          \\ 
VAE-L
$(\frac{\partial\Omega}{\partial\mathcal{\theta}})$                              & 56.18          & 63.94          & 83.04          \\ 
Proposed
& 62.67          & \textbf{64.76} & \textbf{87.57} \\ \hline\hline
\multirow{2}{*}{Method}            & \multicolumn{3}{c}{Blur Types}        \\ \cline{2-4} 
                                   & LB             & DL             & RA             \\ \hline
VAE-A $(z_g)$ & 93.49          & 93.70          & 94.10          \\ 
VAE-R
$(\frac{\partial\mathcal{L}}{\partial\mathcal{\phi}})$                                  & 89.08          & 90.83          & 92.57          \\ 
VAE-L
$(\frac{\partial\Omega}{\partial\mathcal{\theta}})$                              & 94.51          & 94.08          & 94.74          \\ 

Proposed
& \textbf{95.26}          & \textbf{95.71} & \textbf{95.85  }        \\ \hline\hline
\end{tabular}
\vspace{-4mm}
\end{table}
\
In this section, we show the advantages of the gradient analysis framework on the visual tasks detailed in Section~\ref{sec:Experiments} : perceptual image quality assessment and out-of-distribution classification.
\vspace{-1.5mm}
\subsection{Image Quality Assessment}
\vspace{-1.5mm}
The proposed method detailed in Section~\ref{subsec:IQA} is compared against $17$ other state-of-the-art IQA approaches in Table~\ref{tab:tab_results_all}. The performance of the quality estimators is validated using root mean square error (accuracy), outlier ratio (consistency), Pearson correlation (linearity), and Spearman correlation (monotonic behavior). In Table~\ref{tab:tab_results_all}, higher Spearman and Pearson correlations, and lower RMSE and OR are desired. Statistical significance between correlation coefficients is measured with the formulations suggested in ITU-T Rec. P.1401~\cite{ITU2012}. In the performance comparison, we use full-reference methods based on perceptually-extended fidelity~\cite{ponomarenko2011modified}, structural similarity~\cite{wang2004image, wang2003multiscale}, spectral similarity~\cite{zhang2011fsim}, perceptual similarity~\cite{temel2015persim, Temel201692}, SAE based activations~\cite{temel2016unique}, error spectrum analysis~\cite{Temel_SPIC_2018} and no-reference methods based on natural scene statistics~\cite{mittal2012no, moorthy2010two, saad2012blind}.

In each evaluation metric and both databases, we highlight the results of two best performing methods. Statistical significance results with respect to correlation values of the proposed method are reported against the correlation values of other methods. A $0$ corresponds to statistically similar performance, $-1$ means compared method is statistically inferior, and $1$ indicates compared method is statistically superior. None of the compared methods are statistically superior to the proposed method while all of them are statistically inferior to the proposed method in at least one evaluation metric. Note that the proposed gradient based method outperforms the activation based SAE method in~\cite{temel2016unique} in all categories and in both databases. This comparison is an indicator that gradient based projections provide better characterization for distorted representation spaces compared to activations.

\vspace{-1.5mm}
\subsection{Out-Of-Distribution Classification}
\vspace{-1.5mm}
The results of the proposed gradient based method detailed in Section~\ref{subsec:OOD-C} for out-of-distribution classification is shown in Table~\ref{tab:OOD}. Classification accuracy is used as the performance metric to compare results. VAE-A is a method that the activations, $z_g$, of test images are used as features to train the classifier. VAE-R and VAE-L refer to the methods that utilize regularization and reconstruction loss gradients as features, respectively. Proposed is the method that utilizes both regularization and reconstruction gradients. The out-of-distribution challenge types are divided into two categories: spatial blur vs others. The spatial blur category includes Lens Blur, Dirty Lens, and Rain. These challenges decrease the higher frequency components in the test images and hence are categorized together. The other challenge category includes Decolorization, Codec Error, and Noise. The method with the highest classification accuracy is highlighted.

Table~\ref{tab:OOD} demonstrates that gradient features from the VAE are better indicators of being outside the trained representation space than activations taken from the same VAE. Moreover, the classification accuracies based on different gradient features are similar for challenges in the blur category while their accuracy varies significantly among challenges in the other category. This is because the directional information provided by reconstruction loss (VAE-L) and regularization loss (VAE-R) are both equally haphazard for blurred images. This is in contrast to in-distribution (challenge-free) images which all have similar directional gradients. Hence, it is easy to classify between in-distribution and out-of-distribution images using any gradients. On the other hand, in the other non-blur type category, there is considerable variation in accuracy from different gradient features which merits further investigation. 
\vspace{-1.5mm}

\vspace{-1.5mm}

\section{Conclusion}
\vspace{-1.5mm}
In this paper, we proposed a framework to analyze the backpropagated gradients of representation space and show that they are effective directional measures when analyzing distorted images. We motivated the use of gradients by geometric interpretations and proposed a methodology to extract gradients. We conducted experiments that demonstrated the use of gradients as both features as well as error projection spaces in two applications. We also analyzed the effect of regularization on the proposed method. Finally, in both applications, the proposed method using gradients outperformed the features obtained from activations in every measure.
\vspace{-0.5cm}
\bibliographystyle{IEEEbib.bst}
\bibliography{references}

\begin{thebibliography}{10}

\bibitem{Goodfellow2016}
I.~Goodfellow, Y.~Bengio, and A.~Courville,
\newblock {\em Deep Learning},
\newblock MIT Press, 2016,
\newblock \url{www.deeplearningbook.org}.

\bibitem{Bengio2013}
Y.~Bengio, A.~Courville, and P.~Vincent,
\newblock ``Representation learning: A review and new perspectives,''
\newblock {\em IEEE Trans. Patt. Analy. Mach. Intel.}, vol. 35, no. 8, pp.
  1798--1828, 2013.

\bibitem{krizhevsky2012imagenet}
A.~Krizhevsky, I.~Sutskever, and G.~E. Hinton,
\newblock ``Imagenet classification with deep convolutional neural networks,''
\newblock in {\em Neur. Info. Proc. Syst.}, 2012, pp. 1097--1105.

\bibitem{stockham1972image}
T.~G. Stockham,
\newblock ``Image processing in the context of a visual model,''
\newblock {\em Proc. IEEE}, vol. 60, no. 7, pp. 828--842, 1972.

\bibitem{rumelhart1986learning}
D.~E. Rumelhart, G.~E. Hinton, and R.~J. Williams,
\newblock ``Learning representations by back-propagating errors,''
\newblock {\em Nature}, vol. 323, no. 6088, pp. 533, 1986.

\bibitem{patel2015visual}
V.~M. Patel, R.~Gopalan, R.~Li, and R.~Chellappa,
\newblock ``Visual domain adaptation: A survey of recent advances,''
\newblock {\em IEEE Sig. Proc. Mag.}, vol. 32, no. 3, pp. 53--69, 2015.

\bibitem{ng2011sparse}
A.~Ng,
\newblock ``Sparse autoencoder,''
\newblock {\em CS294A Lecture notes}, vol. 72, no. 2011, pp. 1--19, 2011.

\bibitem{Kingma2013}
D.~P. Kingma and M.~Welling,
\newblock ``Auto-encoding variational bayes,''
\newblock {\em arXiv:1312.6114}, 2013.

\bibitem{zou2005regularization}
H.~Zou and T.~Hastie,
\newblock ``Regularization and variable selection via the elastic net,''
\newblock {\em Jour. Royal Stat. Soc.: Ser. B}, vol. 67, no. 2, pp. 301--320,
  2005.

\bibitem{ng2012ufldl}
A.~Ng, J.~Ngiam, C.~Y. Foo, Y.~Mai, and Caroline S.,
\newblock ``Ufldl tutorial,'' 2012.

\bibitem{selvaraju2017grad}
R.~R. Selvaraju, M.~Cogswell, A.~Das, R.~Vedantam, D.~Parikh, and D.~Batra,
\newblock ``Grad-cam: Visual explanations from deep networks via gradient-based
  localization,''
\newblock in {\em IEEE Int. Conf. Comp. Vis.}, 2017, pp. 618--626.

\bibitem{jayaraman2012objective}
D.~Jayaraman, A.~Mittal, A.~K. Moorthy, and A.~C. Bovik,
\newblock ``Objective quality assessment of multiply distorted images,''
\newblock in {\em Asilomar Conf. Sig. Syst. Comp.}, 2012, pp. 1693--1697.

\bibitem{ponomarenko2015image}
N.~Ponomarenko, L.~Jin, O.~Ieremeiev, V.~Lukin, K.~Egiazarian, J.~Astola,
  Benoit Vozel, K.~Chehdi, M.~Carli, F.~Battisti, et~al.,
\newblock ``Image database tid2013: Peculiarities, results and perspectives,''
\newblock {\em Sig. Proc.: Imag. Comm.}, vol. 30, pp. 57--77, 2015.

\bibitem{moorthy2010two}
A.~K. Moorthy and A.~C. Bovik,
\newblock ``A two-step framework for constructing blind image quality
  indices,''
\newblock {\em IEEE Sig. Proc. Let.}, vol. 17, no. 5, pp. 513--516, 2010.

\bibitem{saad2012blind}
M.~A. Saad, A.~C. Bovik, and C.~Charrier,
\newblock ``Blind image quality assessment: A natural scene statistics approach
  in the dct domain,''
\newblock {\em IEEE Trans. on Image Proc.}, vol. 21, no. 8, pp. 3339--3352,
  2012.

\bibitem{kang2014convolutional}
L.~Kang, P.~Ye, Y.~Li, and D.~Doermann,
\newblock ``Convolutional neural networks for no-reference image quality
  assessment,''
\newblock in {\em IEEE Conf. Comp. Vis. Patt. Recog.}, 2014, pp. 1733--1740.

\bibitem{temel2016unique}
D.~Temel, M.~Prabhushankar, and G.~AlRegib,
\newblock ``{UNIQUE}: Unsupervised image quality estimation,''
\newblock {\em IEEE Sig. Proc. Let.}, vol. 23, no. 10, pp. 1414--1418, 2016.

\bibitem{Prabhushankar2017_EI}
M.~Prabhushankar, D.~Temel, and G.~AlRegib,
\newblock ``{MS-UNIQUE}: Multi-model and sharpness-weighted unsupervised image
  quality estimation,''
\newblock {\em Elect. Imag.}, vol. 2017, no. 12, pp. 30--35, 2017.

\bibitem{Liang2017}
S.~Liang, Y.~Li, and R.~Srikant,
\newblock ``Enhancing the reliability of out-of-distribution image detection in
  neural networks,''
\newblock {\em arXiv:1706.02690}, 2017.

\bibitem{Zong2018}
B.~Zong, Q.~Song, M.~R. Min, W.~Cheng, et~al.,
\newblock ``Deep autoencoding gaussian mixture model for unsupervised anomaly
  detection,''
\newblock {\em Int. Conf. Learning Repr.}, 2018.

\bibitem{Fan2018}
Y.~Fan, G.~Wen, D.~Li, S.~Qiu, and M.~D. Levine,
\newblock ``Video anomaly detection and localization via gaussian mixture fully
  convolutional variational autoencoder,''
\newblock {\em arXiv:1805.11223}, 2018.

\bibitem{Temel2017}
D.~Temel, G.~Kwon*, M.~Prabhushankar*, and G.~AlRegib,
\newblock ``{CURE-TSR: Challenging unreal and real environments for traffic
  sign recognition},''
\newblock in {\em Neur. Info. Proc. Syst. Work. on MLITS}, Long Beach, U.S.,
  December 2017.

\bibitem{Temel2018_SPM}
D.~{Temel} and G.~{AlRegib},
\newblock ``Traffic signs in the wild: Highlights from the ieee video and image
  processing cup 2017 student competition [sp competitions],''
\newblock {\em IEEE Sig. Proc. Mag.}, vol. 35, no. 2, pp. 154--161, March 2018.

\bibitem{Temel2019_VIP}
D.~Temel, T.~Alshawi, M-H. Chen, and G.~AlRegib,
\newblock ``Challenging environments for traffic sign detection: Reliability
  assessment under inclement conditions,''
\newblock {\em arXiv:1902.06857}, 2019.

\bibitem{ponomarenko2011modified}
N.~Ponomarenko, O.~Ieremeiev, V.~Lukin, K.~Egiazarian, and M.~Carli,
\newblock ``Modified image visual quality metrics for contrast change and mean
  shift accounting,''
\newblock in {\em Proc. CADSM}, 2011, pp. 305--311.

\bibitem{wang2004image}
Z.~Wang, A.~C Bovik, H.~R. Sheikh, and E.~P. Simoncelli,
\newblock ``Image quality assessment: from error visibility to structural
  similarity,''
\newblock {\em IEEE Trans. Image Proc.}, vol. 13, no. 4, pp. 600--612, 2004.

\bibitem{wang2003multiscale}
Z.~Wang, E.~P Simoncelli, and A.~C. Bovik,
\newblock ``Multiscale structural similarity for image quality assessment,''
\newblock in {\em Asilomar Conf. Sig., Syst. \& Comp.}, 2003, vol.~2, pp.
  1398--1402.

\bibitem{sampat2009complex}
M.~P. Sampat, Z.~Wang, S.~Gupta, A.~C. Bovik, and M.~K. Markey,
\newblock ``Complex wavelet structural similarity: A new image similarity
  index,''
\newblock {\em IEEE Trans. Image Proc.}, vol. 18, no. 11, pp. 2385--2401, 2009.

\bibitem{wang2011information}
Z.~Wang and Q.~Li,
\newblock ``Information content weighting for perceptual image quality
  assessment,''
\newblock {\em IEEE Trans. Image Proc.}, vol. 20, no. 5, pp. 1185--1198, 2011.

\bibitem{zhang2012sr}
L.~Zhang and H.~Li,
\newblock ``{SR-SIM}: A fast and high performance iqa index based on spectral
  residual,''
\newblock in {\em IEEE Int. Conf. Image Proc.}, 2012, pp. 1473--1476.

\bibitem{zhang2011fsim}
L.~Zhang, L.~Zhang, X.~Mou, D.~Zhang, et~al.,
\newblock ``Fsim: a feature similarity index for image quality assessment,''
\newblock {\em IEEE Trans. Image Proc.}, vol. 20, no. 8, pp. 2378--2386, 2011.

\bibitem{mittal2012no}
A.~Mittal, A.~K. Moorthy, and A.~C. Bovik,
\newblock ``No-reference image quality assessment in the spatial domain,''
\newblock {\em IEEE Trans. Image Proc.}, vol. 21, no. 12, pp. 4695--4708, 2012.

\bibitem{temel2015persim}
D.~Temel and G.~AlRegib,
\newblock ``{PerSIM}: Multi-resolution image quality assessment in the
  perceptually uniform color domain,''
\newblock in {\em IEEE Int. Conf. Image Proc.}, 2015, pp. 1682--1686.

\bibitem{Temel201692}
D.~Temel and G.~AlRegib,
\newblock ``{CSV}: Image quality assessment based on color, structure, and
  visual system,''
\newblock {\em Sig. Proc.: Image Comm.}, vol. 48, pp. 92 -- 103, 2016.

\bibitem{Temel_SPIC_2018}
D.~Temel and G.~AlRegib,
\newblock ``Perceptual image quality assessment through spectral analysis of
  error representations,''
\newblock {\em Sig. Proc.: Image Comm.}, 2019.

\bibitem{ITU2012}
ITU-T,
\newblock ``P.1401: Methods, metrics and procedures for statistical evaluation,
  qualification and comparison of objective quality prediction models,''
\newblock Tech. {R}ep., ITU Telecom. Stand. Sector, 2012.

\end{thebibliography}

\end{document}